\documentclass[journal]{IEEEtran}
\usepackage{authblk}
\usepackage[utf8]{inputenc}
\usepackage[top=80pt,bottom=80pt,left=88pt,right=86pt]{geometry}
\usepackage{graphicx}
\usepackage{amsmath,amsfonts}
\usepackage[position=top,labelformat=empty,caption=false,font=small,labelfont=bf,textfont=bf]{subfig}
\usepackage{textcomp}
\usepackage{stfloats}
\DeclareMathOperator*{\argmax}{argmax}
\usepackage{url}   

\usepackage{algorithm}
\usepackage[noend]{algpseudocode}
\usepackage{setspace}
\usepackage{multirow}
\let\Algorithm\algorithm
\renewcommand\algorithm[1][]{\Algorithm[#1]\setstretch{1.1}}

\algrenewcommand{\algorithmiccomment}[1]{\hskip3px$\#$ #1}
\algdef{SE}[SUBALG]{Indent}{EndIndent}{}{\algorithmicend\ }%
\algtext*{Indent}
\algtext*{EndIndent}
\usepackage{balance}
\usepackage{multicol}

\begin{document}

\title{\huge An Evolutionary, Gradient-Free, Query-Efficient, Black-Box Algorithm for Generating Adversarial Instances in Deep Networks}
\author{Raz Lapid, Zvika Haramaty, Moshe Sipper
\thanks{The authors are with the Department of Computer Science, Ben-Gurion University, Beer Sheva 84105, Israel. Corresponding author: R. Lapid, razla@post.bgu.ac.il.}}
\date{\today}

\markboth{Lapid et al.}{Lapid et al.}

\maketitle

\begin{abstract}
Deep neural networks (DNNs) are sensitive to adversarial data in a variety of scenarios, including the black-box scenario, where the attacker is only allowed to query the trained model and receive an output. Existing black-box methods for creating adversarial instances are costly, often using gradient estimation or training a replacement network. This paper introduces \textbf{Qu}ery-Efficient \textbf{E}volutiona\textbf{ry} \textbf{Attack}, \textit{QuEry Attack}, an untargeted, score-based, black-box attack. QuEry Attack is based on a novel objective function that can be used in gradient-free optimization problems. The attack only requires access to the output logits of the classifier and is thus not affected by gradient masking. No additional information is needed, rendering our method more suitable to real-life situations. We test its performance with three different state-of-the-art models---Inception-v3, ResNet-50, and VGG-16-BN---against three benchmark datasets: MNIST, CIFAR10 and ImageNet. Furthermore, we evaluate QuEry Attack's performance on non-differential transformation defenses and state-of-the-art robust models. Our results demonstrate the superior performance of QuEry Attack, both in terms of accuracy score and query efficiency.
\end{abstract}


\section{Introduction}
\IEEEPARstart{D}{eep} neural networks (DNNs) have become the central approach in modern-day artificial intelligence (AI) research. They have attained superb performance in multifarious complex tasks and are behind fundamental breakthroughs in a variety of machine-learning tasks that were previously thought to be too difficult. Image classification, object detection, machine translation, and sentiment analysis are just a few examples of domains revolutionized by DNNs.

Despite their success, recent studies have shown that DNNs are vulnerable to adversarial attacks. A barely detectable change in an image, for example, can cause a misclassification in a well-trained DNN. Targeted adversarial examples can even evoke a misclassification of a specific class (e.g., misclassify a car as a cat). Researchers have demonstrated that adversarial attacks are successful in the real world and may be produced for data modalities beyond imaging, e.g., natural language and voice recognition \cite{wang2019natural,morris2020textattack,carlini2018audio,schonherr2018adversarial}. DNNs' vulnerability to adversarial attacks has raised concerns about applying these techniques to safety-critical applications.

To discover effective adversarial instances, most past work on adversarial attacks has employed gradient-based optimization \cite{goodfellow2014explaining,papernot2016limitations,carlini2017towards,gu2014towards,moosavi2016deepfool}. Gradient computation can only be executed if the attacker is fully aware of the model architecture and weights. Thus, these approaches are only useful in a white-box scenario, where an attacker has complete access and control over a targeted DNN. Attacking real-world AI systems, however, might be far more arduous. The attacker must consider the difficulty of implementing adversarial instances in a black-box setting, in which no information about the network design, parameters, or training data is provided. In this situation, the attacker is exposed only to the classifier's input-output pairs. In this context, a typical strategy has been to attack trained replacement networks and hope that the generated examples transfer to the target model \cite{papernot2017practical}. The substantial mismatch of the model between the alternative model and the target model, as well as the significant computational cost of alternative network training, often renders this technique ineffective.

In our work we assume a real-world, black-box attack scenario, wherein a DNN's input and output may be accessed but not its internal configuration. We focus on a scenario in which a specific DNN is an image classifier, specifically, a convolutional neural network (CNN), which accepts an image as input and outputs a probability score for each class. 

Herein, we present an evolutionary, gradient-free optimization approach for generating adversarial instances. Our proposed attack can deal with either constrained ($\epsilon$ value that constrains the norm of the allowed perturbation) or unconstrained (no constraint on the norm of the perturbation) problems, and focuses on constrained, untargeted attacks. We believe that our framework can be easily adapted to the targeted setting.

In the next section we review the literature on adversarial attacks. Section~\ref{sec:threat_model} summarizes the threat model we assume for our proposed evolutionary attack algorithm. The algorithm itself---\textit{QuEry Attack} (for \textbf{Qu}ery-Efficient \textbf{E}volutiona\textbf{ry} \textbf{Attack})---is delineated in Section~\ref{sec:query}. The experiments conducted to test the method, along with results, are described in Section~\ref{sec:experiments}. We discuss our findings and present concluding remarks in Section~\ref{sec:conclusions}.

\begin{figure}
\begin{tabular}{ccc}
Original image & Successful attack & Failed attack \\[5pt]
\includegraphics[width=0.13\textwidth]{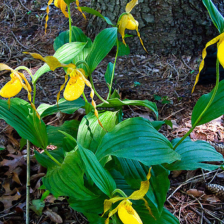} & 
\includegraphics[width=0.13\textwidth]{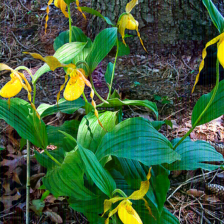} &
\includegraphics[width=0.13\textwidth]{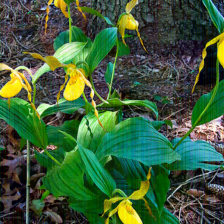} \\[5pt]

\includegraphics[width=0.13\textwidth]{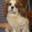} &
\includegraphics[width=0.13\textwidth]{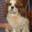} &
\includegraphics[width=0.13\textwidth]{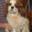} \\[5pt]

\includegraphics[width=0.13\textwidth]{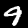} &
\includegraphics[width=0.13\textwidth]{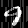} &
\includegraphics[width=0.13\textwidth]{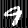} \\
\end{tabular}
\caption{Examples of adversarial attacks generated by QuEry Attack.
         Top row: Imagenet ($l_\infty = 6/255$). 
         Middle row: CIFAR10 ($l_\infty = 6/255$).
         Bottom row: MNIST ($l_\infty = 60/255$).
         Left: the original image.
         Middle: a successful attack.
         Right: A failed attack.}
\label{fig:attacks}
\end{figure}

Figure~\ref{fig:attacks} shows examples of successful and unsuccessful instances of images generated by QuEry Attack, evaluated against ImageNet, CIFAR10, and MNIST.

\section{Related Work}
Adversarial attacks against DNNs have become an important research field in the last few years. For a comprehensive survey, we refer the reader to \cite{Xu2020}.

An important distinction is between `white box' and `black box' attacks. In white-box attacks, the attacker has knowledge of the attacked model's internal structure and parameters, and exploits that knowledge. It is commonly done by using the model's gradients \cite{goodfellow2014explaining,carlini2017towards,computers9030058}.

Although white-box methods achieved good results they usually do not represent a real-world scenario. More realistic is a black-box attack, where the attacker has no knowledge of the model's structure and parameters. The attacker can only query the model---and act upon its outputs. We can further distinguish between the more common `light black box' attacks, where the model gives the prediction's confidence (which can be exploited), and `dark black box' attacks where the model only gives the final class prediction (e.g., \cite{papernot2017practical}).

The first effective black-box attack traded a-priori information of the model with extensive runtime querying \cite{Papernot2017}. Using a large number of queries it builds a substitute model, and attacks it with traditional white-box methods. It uses the transferability property, namely, an attack that succeeds on one model will likely succeed on a similar---though not identical---model.

Other works used the more permissive `light black box' scenario, which can use the prediction's confidence value. Some works estimate the gradient with this information and then use traditional gradient-based attacks \cite{Chen2017}.

All these black-box methods rely on gradients, and thus are sensitive to many defense methods that obscure gradients \cite{buckman2018thermometer,guo2017countering,dhillon2018stochastic}. This has given rise to methods that do not rely on gradients at all, e.g., \cite{andriushchenko2020square}, which uses random search and is also query-efficient.

Instead of randomness, one can use evolutionary methods. 
In \textit{Evolutionary Algorithms} (EAs), core concepts from evolutionary biology—--inheritance, random variation, and selection--—are harnessed in algorithms that are applied to complex computational problems. EA techniques have been shown to solve numerous difficult problems from widely diverse domains, and also to produce human-competitive machine intelligence \cite{Kannappan2015}.
EAs also present many important benefits over popular Machine Learning methods, including \cite{Sipper2017ec}: less reliance on the existence of a known or discoverable gradient within the search space; ability to handle design problems, where the objective is to design new entities from scratch; fewer required a priori assumptions about the problem at hand; seamless integration of human expert knowledge; ability to solve problems where human expertise is very limited; support of interpretable solution representations; support of multiple objectives. 

The evolutionary method GenAttack is a \textit{targeted} attack (thus not directly comparable to ours, which is \textit{untargeted}) that used a fitness function that tries to increase the probability of the target class and decrease the probability of the other classes \cite{10.1145/3321707.3321749}. Its fitness function ignored the distance between the images. Interestingly, GenAttack uses fitness-proportionate selection, which is employed less often nowadays due to known problems. It uses an adaptive mutation rate to balance between exploration in early phases and exploitation in later phases.

\cite{Prochazka2020} treated the adversarial problem as one of multi-objective optimization: minimize the class prediction's score on one hand, and the distance between the original image and a perturbed one on the other hand.

Another attack method changes a single pixel \cite{Su2019}. This method uses differential evolution (DE) \cite{Das2011a} without crossover. However, it sometimes required thousands of queries.

\cite{Lin2020} also uses DE, but unlike the other evolutionary computation (EC) methods reviewed, it uses EC to approximate the gradients.

Unlike the above methods, which tried to minimize the perturbation as much as possible and make it as unnoticeable to the human eye as possible, \cite{Jere2019a} makes a small but noticeable change, which looks like a regular scratch (a similar approach in the domain of Natural Language Processing creates sentences that do not make sense to a human reader \cite{DiGiovanni2021}).
The approach uses DE as well, and also Covariance-Matrix Adaptation Evolution Strategies (CMA-ES) \cite{hansen2001completely}. Unlike most attacks, which use the $l_{\infty}$ or $l_2$ norms, this one is based on the $l_0$ norm.

In the EC methods seen so far, evolution is run against a \emph{single} image,
and each individual is a perturbation added to that image. \cite{Wang2018}, on the other hand, used the transferability property mentioned earlier to evolve a universal perturbation.
An individual is an image mask that can be applied as a perturbation to any image.

Our extensive scrutiny of the literature and software repositories revealed that many authors compare their work to prior works that do not use the same threat model: There might be a mismatch in norms (e.g., $l_2$ vs. $l_{\infty}$), white box vs. black box, or other subtle differences. 
Moreover, having investigated numerous software repositories, we found that running the code of many papers is far from straightforward. 
%Some works are not accompanied by published code. Others do make published code available, albeit lacking basic instructions and documentation, rendering it arduous to run the code for comparative purposes. For some works we eventually succeeded in executing the published code, which, alas did not report any results; it was thus unclear to us how to make use of such code. We reached out to some of the original authors, but somewhat disappointingly received almost no response. The comparison setup we delineate in Section~\ref{sec:experiments} was thus obtained with great effort and the attacks and defenses were carefully selected.

\section{Threat Model}
\label{sec:threat_model}
In the black-box attack setting, queries to the network are permitted but access to internal states is prohibited (e.g., executing backpropagation). Hence, our threat model, or scenario, is as follows:
\begin{itemize}
    \item The attacker is unaware of the network's design, settings, or training data.
    \item The attacker has no access to intermediate values in the target model.
    \item The attacker can only use a black-box function to query the target model.
\end{itemize}

Note that the above threat model determines the comparisons we perform, which focus on attacks that are:
\begin{enumerate}
    \item black-box,
    \item untargeted, 
    \item $l_{\infty}$-norm bounded.
\end{enumerate}

We can consider a network model to be a function:
\begin{equation}
f: [0,1]^d \rightarrow \mathbb{R}^C,
\end{equation}
where $d$ is the number of input features and $C$ is the number of classes. The $c$-th value $f_c(x) \in \mathbb{R}$ specifies the predicted score of classifying input image $x$ as class $c$. The classifier assigns class $y=\argmax_{c=1,...,C}f_c(x)$ to the input $x$. 

A \textit{targeted} attack aims to create an image that will be incorrectly classified into a given (incorrect) class. An \textit{untargeted} attack aims to create an image that will be incorrectly classified into \textit{any} class except the correct one. An image $\hat{x}$ is termed an adversarial example, with an $l_p$-norm bound of $\epsilon$ for $x$, if:

\begin{gather}
    \nonumber
    \argmax_{c=1,...,C}f_c(\hat{x}) \neq y,
    \\
    s.t. \ \Vert \hat{x} - x \Vert_p \leq \epsilon \ \text{and} \ \hat{x}\in [0,1]^d.
\end{gather}

To wit, the model should classify $\hat{x}$ incorrectly, while preserving similarity between $x$ and $\hat{x}$ under an $l_p$ distance metric.

We focus on a black-box, score-based attack, wherein the only information of the threat model is the raw output (logits). 

Our suggested black-box approach may theoretically be used in conjunction with classic machine-learning models, with the same input-output relationship. Because DNNs have reached state-of-the-art performance in a variety of image tasks, we focus on them in this paper.

\section{QuEry Attack}
\label{sec:query}
QuEry Attack is an evolutionary algorithm (EA) that explores a space of images, defined by a given input image and a given input model, in search of adversarial instances. 
It ultimately generates an attacking image for the given input image.
Unlike white-box approaches, we make no assumptions about the targeted model, its architecture, dataset, or training procedure. We assume that we have an image $x$, which a black-box neural network, $f$, classifies by outputting a probability distribution over the set of classes, as stated in Section~\ref{sec:threat_model}. The actual label $y$ is computed as $y=\argmax f(x)$. 

Our objective is to find a perturbed image, $\hat{x}$, of image $x$, such that, $\Vert \hat{x} - x \Vert_{\infty} \leq \epsilon$, which causes the network to predict $y'=\argmax f(\hat{x})$, such that $y' \neq y$. Finding $\hat{x}$ may be cast as a constrained optimization problem:
\begin{equation}
\label{eq:loss}
    \min_{\hat{x}\in[0,1]^d}{\mathcal{L}(f(\hat{x}), y)}, \ \ s.t. \ \ \Vert \hat{x} - x \Vert_{\infty} \leq \epsilon,
\end{equation}
for a given loss function $\mathcal{L}$. 

We use loss $\mathcal{L}$ as the fitness function, defined in our case as:
\begin{equation}
\label{eq:fitness}
    \mathit{fitness}(\hat{x})= f_y(\hat{x}) - \max_{c\neq y}f_c(\hat{x}) + \lambda \Vert \hat{x} - x \Vert_{2},
\end{equation}
where $\hat{x}$ is a perturbed image, $f_y$ is the predicted score that $\hat{x}$ belongs to class $y$, $f_c$ is the predicted score that $\hat{x}$ belongs to class $c\neq y$.
In order to guarantee that the adversarial perturbation is as imperceptible as possible we penalize the $l_2$ distortion of the perturbation by including a regularization component in the fitness function. We use $l_2$ regularization because we noticed that most of the evolved adversarial examples were on the edges of the $\epsilon$-ball, and we wanted to give precedence to examples which were closer to the original input. This penalization is determined by the $\lambda$ value, which is the regularization strength. In our experiments we used $\lambda=1$.

The ultimate goal is to minimize the fitness value:
Essentially, the lower the logit of the correct class and the higher the maximum logit of the other classes---the better the value.

Algorithm~\ref{alg:query} provides the pseudo-code of QuEry Attack. The original image $x$, along with a number of hyperparameters, are given as input to the algorithm. QuEry Attack generates an adversarial image $\hat{x}$, with the model classifying $\hat{x}$ as $y'$ such that $y' \neq y$ and $\Vert \hat{x} - x \Vert_{\infty} \leq \epsilon$.

The main goal of QuEry Attack is to produce a successful attack, using as few queries to the model as possible. The maximal number of queries equals generation count $(G)$ $\times$ population size $(N)$.

\begin{algorithm*}
\footnotesize
\caption{QuEry Attack}
\label{alg:query}
\begin{multicols}{2}
\begin{algorithmic}[1]
\Require
\Indent
\Statex $x$ $\gets$ original image
\Statex $y$ $\gets$ original label
\Statex $\epsilon$ $\gets$ maximum $l_\infty$ distance
\Statex $p$ $\gets$ proportion of elements in $x$ to be perturbed
\Statex $N$ $\gets$ population size
\Statex $G$ $\gets$ maximum number of generations
\Statex $T$ $\gets$ tournament size
\EndIndent

\Statex

\Ensure
\Indent
\Statex $\hat{x}$ $\gets$ adversarial image
\EndIndent

\Statex
\Statex
\noindent\hskip-\leftmargin \Comment{Main loop}
\State $gen \gets 0$
\State $pop \gets$ \textsc{init}()
\While{not \textsc{termination\_condition($pop, gen$)}}
   \For{$\hat{x} \in pop$} 
        \State compute fitness of $\hat{x}$ using Equation~\ref{eq:fitness}
   \EndFor
   \State $new\_pop \gets \emptyset$
   \State $elite \gets$ \textsc{elitism($pop$)}
   \State add $elite$ to $new\_pop$
   \For{$i \gets 1$ to $\frac{P - 1}{3}$}
        \State $parent_1 \gets$ \textsc{selection($pop$)}
        \State $parent_2 \gets$ \textsc{selection($pop$)}
        \State $\mathit{offspring_1, offspring_2} \gets$ \textsc{crossover($pop$)}
        \State $\mathit{mut_1} \gets$ \textsc{square\_mutation($\mathit{offspring_1}$)}
        \State $\mathit{mut_2} \gets$ \textsc{square\_mutation($\mathit{offspring_2}$)}
        \State add $\mathit{offspring_1, mut_1, mut_2}$ to $new\_pop$ 
   \EndFor
   \State $pop \gets new\_pop$
   \State $gen \gets gen + 1$
\EndWhile
\State
\State \textbf{return} best $\hat{x}$ from $pop$ \Comment{QuEry Attack's final output}

\Statex
\Function{init}{\,}
    \State $pop \gets \emptyset$
    \For{$i$ $\gets$ 1 to $N$} 
        \State $\hat{x} \gets$  \textsc{stripes\_init}(\textit{x})
        \State add $\hat{x}$ to $pop$
    \EndFor
    \State \textbf{return} $pop$
\EndFunction

\Statex
\Function{elitism}{\textit{pop}}
    \State \textbf{return} best $\hat{x}$ from $pop$
\EndFunction

\Statex
\Function{termination\_condition}{\textit{pop, gen}}
    \If{$gen = G$}
        \State \textbf{return} true
    \EndIf
    \For{$\hat{x} \in pop$}
        \State $\hat{y} \gets$ predicted label of $\hat{x}$
        \If{$\hat{y} \neq y$}
            \State \textbf{return} true
        \EndIf
    \EndFor
    \State \textbf{return} false
\EndFunction

\Statex
\Function{selection}{\textit{pop}}
    \State $tournament \gets $ randomly and uniformly pick $T$ individuals from $pop$
    \State \textbf{return} best $\hat{x}$ from $tournament$
\EndFunction

\Statex
\Function{stripes\_init}{$\hat{x}$}
    \For{$i$ $\gets$ 1 to $c$} \Comment{c is the image's number of channels}
        \State \textit{stripe} $\gets$ create a vertical stripe of width 1, randomly positioned, with random values $\in \{-\epsilon , \epsilon \}$
        \State $\hat{x}$ $\gets$ $\hat{x}$ + \textit{stripe}
    \EndFor
    \State $\hat{x}$ $\gets \Pi_\epsilon(\hat{x})$ \Comment{$\Pi_\epsilon$: clipping operator to ensure pixel values are within $\epsilon$-ball}
    \State \textbf{return} $\hat{x}$
\EndFunction

\Statex
\Function{square\_mutation}{$\hat{x}$}
    \State $c \gets$ number of channels
    \State $f \gets$ number of features ($h \times w)$ \Comment{$h$: height, $w$: width}
    \State $k \gets \lceil \sqrt{p \cdot f} \rceil$
    \State $\delta \gets $ array of ones of size $k \times k \times c$.
    \State $row, col \gets \mathcal{U}(\{0,...,w-h\})$ \Comment{$\mathcal{U}$ randomly and uniformly selects from given set}
    \For{$i$ $\gets$ 1 to $c$}
            \State $\tau$ $\gets$ $\mathcal{U}(\{-2\epsilon, 2\epsilon\})$
            \State $\delta_{row+1:row+h, col+1:col+h, i} \gets \tau \cdot \delta$
            \State $\hat{x} \gets \hat{x} + \delta$
    \EndFor
    \State $\hat{x} \gets \Pi_\epsilon(\hat{x})$ 
    \State \textbf{return} $\hat{x}$
\EndFunction

\Statex
\Function{crossover}{\textit{$parent_1$}, \textit{$parent_2$}} 
    \State Flatten \textit{$parent_1$} and \textit{$parent_2$}
    \State Perform standard two-point crossover (as explained in text), creating $\mathit{offspring_1, offspring_2}$
    \State $\mathit{offspring_1, offspring_2}$ $\gets \Pi_\epsilon(\mathit{offspring_1}), \Pi_\epsilon(\mathit{offspring_2})$
    \State \textbf{return} $\mathit{offspring_1, offspring_2}$
\EndFunction

\end{algorithmic}
\end{multicols}

\normalsize
\end{algorithm*} 

\paragraph{Initialization}
Initialization is crucial for optimization problems, e.g., in deep-learning training, gradient descent reaches a local minimum that is significantly determined by the initialization technique \cite{kumar2017weight,koturwar2017weight}.
QuEry Attack generates an initial population of perturbed images by randomly selecting images from the edges of the sphere centered on the original image $x$ with radius = $\epsilon$. This is accomplished by adding vertical stripes of width 1 along the image, with the color of each stripe sampled uniformly at random from $\{-\epsilon, \epsilon\}$ per channel (i.e., the pixels of each stripe can be either $-\epsilon$ or  $\epsilon$); in \cite{andriushchenko2020square}, they discovered that convolutional neural networks are especially vulnerable to such perturbations. 

\paragraph{Mutation}
Considering the use of (square-shaped) convolutional filters by convolutional neural networks, we used square-shaped perturbations. Specifically, we employed \cite{andriushchenko2020square}'s technique for determining square size. 
Let $p\in[0,1]$ be the proportion of elements to be perturbed for an image of shape $ h \times w$.
The nearest positive integer to $\sqrt{p \cdot h \cdot w}$ determines the length of the square's edge, with $p$ being a hyperparameter. We set it initially to $p=0.1$, then halved it after $\{40, 200, 800, 4000, 8000, 16000, 24000, 32000\}$ queries, respectively (similar to \cite{andriushchenko2020square}). 

\paragraph{Crossover}
We experimented both with single-point and two-point crossover, eventually settling on the latter as it performed better. The operator works by flattening both (two-dimensional image) parents, randomly picking two indices, then swapping the pixels between the chosen pixels.

The EA then proceeds by evaluating the fitness of each individual, selecting parents, and performing crossover and mutation to generate the next generation.
This latter is obtained by adding one elite individual from the current generation, with all other next-generation individuals derived through crossover and mutation.
The process is repeated until a successful perturbation is found or until the termination condition is met.

A major advantage of QuEry Attack is its amenability to parallelization---due to being evolutionary---in contrast to most other adversarial, iterative (non-evolutionary) attacks in this field. 

\section{Experiments and Results}
\label{sec:experiments}

To evaluate QuEry Attack we set out to collect state-of-the-art algorithms for comparative purposes. A somewhat disconcerting reality we then encountered involved our struggle to find good benchmarks and software for comparison purposes. Sadly, we found ourselves wasting many a day (which, alas, turned into weeks) trying to run buggy software, chasing down broken links, issuing GitHub issues, and so forth. Perhaps this is due in part to the field of adversarial attacks being young. 

% We wanted to test our algorithm against GenAttack, being the only other evolutionary technique we found that has a similar threat model, though GenAttack is a targeted attack and QuEry Attack is untargeted---we came across some comparisons between untargeted and targeted attacks in the literature. However,
% GenAttack was quite cumbersome to deal with: required versions of packages were in contradiction, which we tried to solve for many days using multiple tricks; an issue we opened on the GitHub repository went unanswered; we attempted various translations between TensorFlow (used by GenAttack) and PyTorch (the currently more popular DNN package, used by QuEry Attack). We finally gave up on GenAttack.

Our experimental results are summarized in Table~\ref{tab:results}. The code will be available at \url{github.com/razla}. 

\begin{table*}
\centering
\small
\caption{Experimental results.
         ASR: Attack Success Rate.
         Queries: Number of model queries.
         Each value represents the median of 200 runs (images). 
         Epsilon: $\mathcal{E} \in \{0...255\}$ (8-bits pixel values).
         Top results are marked in boldface.}
\label{tab:results}

\subfloat[ImageNet]{
\begin{tabular}{|r|c|cc|cc|cc|}
\hline
\multirow{2}{*}{Model} & \multirow{2}{*}{$\mathcal{E}$} & \multicolumn{2}{c|}{QuEry Attack} & \multicolumn{2}{c|}{Square} & \multicolumn{2}{c|}{AdversarialPSO  } \\ \cline{3-8} 
                        & & \multicolumn{1}{l|}{ASR} & Queries & \multicolumn{1}{l|}{ASR} & Queries & \multicolumn{1}{c|}{ASR} & Queries \\ \hline

\multirow{4}{*}{Inception-v3} & 24  & \multicolumn{1}{c|}{\bf 100\%} & \textbf{1}  & \multicolumn{1}{c|}{\textbf{100\%}}  & 5 & \multicolumn{1}{c|}{98.5\%} & 51 \\ 
                             & 18 & \multicolumn{1}{c|}{\bf 100\%}   & \textbf{2} & \multicolumn{1}{c|}{\textbf{100\%}}  & 8 & \multicolumn{1}{c|}{98.5\%} & 69 \\ 
                             & 12   & \multicolumn{1}{c|}{\bf 100\%}   & \textbf{22}   & \multicolumn{1}{c|}{95.5\%}  & 32 & \multicolumn{1}{c|}{97\%} & 102  \\
                             & 6   & \multicolumn{1}{c|}{\bf 99.5\%}   & 276   & \multicolumn{1}{c|}{99.0\%}  & \textbf{263} & \multicolumn{1}{c|}{95\%} & 285 \\
                             \hline

\multirow{4}{*}{ResNet-50} & 24  & \multicolumn{1}{c|}{\bf 100\%} & 1  & \multicolumn{1}{c|}{99.5\%}  & 5 & \multicolumn{1}{c|}{98.5\%} & 51  \\ 
                             & 18 & \multicolumn{1}{c|}{\bf 100\%}   & \textbf{1} & \multicolumn{1}{c|}{\textbf{100\%}}  & 5 & \multicolumn{1}{c|}{98.5\%} & 69 \\ 
                             & 12   & \multicolumn{1}{c|}{\bf 100\%}   & \textbf{13}   & \multicolumn{1}{c|}{\textbf{100\%}}  & 26 & \multicolumn{1}{c|}{97\%} & 102  \\ & 6   & \multicolumn{1}{c|}{\bf 100\%}   & \textbf{211}   & \multicolumn{1}{c|}{99.0\%}  & 248 & \multicolumn{1}{c|}{95\%} & 285  \\ \hline

\multirow{4}{*}{VGG-16-BN} & 24  & \multicolumn{1}{c|}{\bf 100\%} & \textbf{1}  & \multicolumn{1}{c|}{\textbf{100\%}}  & 5 & \multicolumn{1}{c|}{98.5\%} & 51 \\ 
                             & 18 & \multicolumn{1}{c|}{\bf 100\%}   & \textbf{1} & \multicolumn{1}{c|}{\textbf{100\%}}  & 5 & \multicolumn{1}{c|}{98.5\%} & 69  \\ 
                             & 12   & \multicolumn{1}{c|}{\bf 100\%}   & \textbf{1}   & \multicolumn{1}{c|}{\textbf{100\%}}  & 5 & \multicolumn{1}{c|}{97\%} & 102  \\ & 6   & \multicolumn{1}{c|}{\bf 100\%}   & \textbf{77}   & \multicolumn{1}{c|}{\textbf{100\%}}  & 86 & \multicolumn{1}{c|}{95\%} & 285 \\ \hline
\end{tabular}
}

\subfloat[CIFAR10]{
\begin{tabular}{|r|c|cc|cc|cc|}
\hline
\multirow{2}{*}{Model} & \multirow{2}{*}{$\mathcal{E}$} & \multicolumn{2}{c|}{QuEry Attack} & \multicolumn{2}{c|}{Square} & \multicolumn{2}{c|}{AdversarialPSO} \\ \cline{3-8} 
                        & & \multicolumn{1}{l|}{ASR} & Queries & \multicolumn{1}{l|}{ASR} & Queries & \multicolumn{1}{c|}{ASR} & Queries  \\ \hline

\multirow{4}{*}{Inception-v3} & 24  & \multicolumn{1}{c|}{\bf 100\%} & \textbf{1}  & \multicolumn{1}{c|}{89.5\%}  & 5 & \multicolumn{1}{c|}{97.5\%} & {31}  \\ 
                             & 18 & \multicolumn{1}{c|}{\bf 100\%}   & \textbf{2} & \multicolumn{1}{c|}{92.5\%}  & 17 & \multicolumn{1}{c|}{96.0\%} & 41 \\ 
                             & 12   & \multicolumn{1}{c|}{\bf 97.5\%}   & \textbf{20}   & \multicolumn{1}{c|}{94.5\%}  & 77 & \multicolumn{1}{c|}{ 95.0\%} & 54  \\ 
                             & 6   & \multicolumn{1}{c|}{91.0\%}  & {428} & \multicolumn{1}{c|}{\textbf{95.5\%}}  & 776 & \multicolumn{1}{c|}{ 94.5\%} & \textbf{223}  \\
                             \hline

\multirow{4}{*}{ResNet-50} & 24  & \multicolumn{1}{c|}{\bf 100\%} & \textbf{2}  & \multicolumn{1}{c|}{92.0\%}  & 8 & \multicolumn{1}{c|}{97.5\%} & 31 \\ 
                             & 18 & \multicolumn{1}{c|}{\bf 100\%}   & \textbf{8} & \multicolumn{1}{c|}{91.0\%}  & 23 & \multicolumn{1}{c|}{96.0\%} & 41 \\ 
                             & 12   & \multicolumn{1}{c|}{\bf 100\%}   & {96}   & \multicolumn{1}{c|}{87.0\%}  & 110 & \multicolumn{1}{c|}{95.0\%} & \textbf{54} \\
                             & 6  & \multicolumn{1}{c|}{\textbf{99.0\%}}  & 565 & \multicolumn{1}{c|}{87.0\%}  & {449} & \multicolumn{1}{c|}{ 94.5\%} & \textbf{223}  \\ \hline

\multirow{4}{*}{VGG-16-BN} & 24  & \multicolumn{1}{c|}{\bf 100\%} & \textbf{1}  & \multicolumn{1}{c|}{89.5\%}  & 5 & \multicolumn{1}{c|}{97.5\%} &  31  \\ 
                             & 18 & \multicolumn{1}{c|}{\bf 99.0\%}   & \textbf{2} & \multicolumn{1}{c|}{87.0\%}  & 14 & \multicolumn{1}{c|}{96.0\%} & 41 \\ 
                             & 12   & \multicolumn{1}{c|}{\bf 98.0\%}   & {60}   & \multicolumn{1}{c|}{87.0\%}  & 86 & \multicolumn{1}{c|}{95.0\%} & \textbf{54} \\
                             & 6  & \multicolumn{1}{c|}{\textbf{95.5\%}}  & {741} & \multicolumn{1}{c|}{88.5\%}  & 890 & \multicolumn{1}{c|}{ 94.5\%} & \textbf{223}  \\
                             \hline
\end{tabular}
}

\subfloat[MNIST]{
\begin{tabular}{|r|c|cc|cc|cc|}
\hline
\multirow{2}{*}{Model} & \multirow{2}{*}{$\mathcal{E}$} & \multicolumn{2}{c|}{QuEry Attack} & \multicolumn{2}{c|}{Square} & \multicolumn{2}{c|}{AdversarialPSO} \\ \cline{3-8} 
                        & & \multicolumn{1}{l|}{ASR} & Queries & \multicolumn{1}{l|}{ASR} & Queries & \multicolumn{1}{c|}{ASR} & Queries  \\ \hline

\multirow{2}{*}{Conv Net} & 80  & \multicolumn{1}{c|}{\textbf{100\%}} & \textbf{5} & \multicolumn{1}{c|}{86.0\%} & 14 & \multicolumn{1}{c|}{76\%} & 2675   \\

                    & 60 & \multicolumn{1}{c|}{93.5\%} & \textbf{72} & \multicolumn{1}{c|}{93.0\%}  & 77 & \multicolumn{1}{c|}{\textbf{99\%}} & 292  \\ 

\hline
\end{tabular}
}

\normalsize
\end{table*}

We evaluated QuEry Attack by executing experiments against three state-of-the-art image classification models---Inception-v3 \cite{szegedy2016rethinking}, ResNet-50 \cite{he2016deep}, and VGG-16-BN \cite{simonyan2014very}---over three image datasets: ImageNet, CIFAR-10, and MNIST. We employed 200 randomly picked and correctly classified images from the test sets.  
For ImageNet, Inception-v3 has an accuracy of 78.8\%, ResNet-50 has an accuracy of 76.1\%, and VGG-16-BN has an accuracy of 73.3\% (these are top-1 accuracy values; for ImageNet, top-5 accuracy values are also sometimes given, which in our case are: Inception-v3 -- 94.4\%, ResNet-50 -- 92.8\%, VGG-16-BN -- 91.5\%).
CIFAR10 accuracy values are: Inception-v3 -- 93.7\%, ResNet-50 -- 93.6\%, and VGG-16-BN -- 94.0\%.
For the MNIST dataset we trained a convolutional neural network (CNN), whose architecture is delineated in Table~\ref{tab:convnet}, which attained 98.9\% accuracy.

All accuracy values are over test images. We used the Adversarial Robustness Toolbox (ART) \cite{nicolae2018adversarial} to evaluate QuEry Attack against other attacks. We restricted all attacking algorithms to a maximum of 42K queries to the model ($N=70$, $G=600$) for MNIST and CIFAR10, and 84K queries ($N=70$, $G=1200$) for ImageNet. 
A query refers to a prediction supplied by the model for a given image.
To make the most of the computational resources we had available we prioritized actual, experimental runs over hyperparameter runs, so hyperparameters were chosen through limited trial and error. In the future we plan to perform a more thorough hyperparameter sweep using Optuna \cite{akiba2019optuna}.
The only hyperparameters we set were the population size $N=70$, tournament size $T=25$, and $p=0.1$; these are used for all experiments reported herein. The number of generations ($G$) was derived from the query budget. 
%Note that the collection of black-box, score-based attacks seems exhaustive, due to the fact that most of them do not have working code, or the results were not good enough for comparison.

AdversarialPSO \cite{Mosli2020} results were obtained by running the code in the GitHub repository referred to in their paper. Due to technical difficulties it was run against the original models that this attack was planned to run against, namely, Inception-v3 for ImageNet, and their own trained networks for CIFAR-10 and MNIST. We duplicated these results in the table for all models. 

\begin{table}
\centering
\caption{CNN used for MNIST.}
\label{tab:convnet}

\begin{tabular}{cc}  
\textbf{Conv Block} & \textbf{Hyperparameters} \\  
\resizebox{0.2\textwidth}{!}{
    \begin{tabular}{|c|c|}
        \hline
        Layer & Layer type \\ 
        \hline
        1 & Convolution \\
        2 & BatchNorm2d \\
        3 & ReLU \\
        \hline
    \end{tabular}} &  % starting rightmost sub table
    
% table 2
    \resizebox{0.2\textwidth}{!}{
    \begin{tabular}{|c|c|}
        \hline
        Hyperparameter & Value \\ 
        \hline
        Epochs & $300$ \\
        Batch size & $64$ \\
        Optimizer & $Adam$ \\
        Learning rate & $0.01$ \\
        Weight decay & $1e-6$ \\
        \hline
    \end{tabular}}
\end{tabular}

\vspace{7pt}

\begin{tabular}{c}
    \textbf{CNN Architecture } \\
     \resizebox{0.45\textwidth}{!}{
        \begin{tabular}{|c|c|c|c|c|}
            \hline
             Layer & Layer type & No. channels & Filter size & Stride \\ 
              \hline
              1 & Conv Block & 32 & $3 \times 3$ & 1 \\
              2 & Max Pooling & N/A & $2 \times 2$ & 2 \\
              3 & Conv Block & 64 & $3 \times 3$ & 1 \\
              4 & Max Pooling & N/A & $2 \times 2$ & 2 \\
              5 & Conv Block & 128 & $3 \times 3$ & 1 \\
              6 & Max Pooling & N/A & $2 \times 2$ & 2 \\
              7 & Dropout $(p=0.5)$ & N/A & N/A & N/A \\
              8 & Fully Connected & 128 & N/A & N/A \\
              9 & Fully Connected & 10 & N/A & N/A \\
              \hline
        \end{tabular}} \\
\end{tabular}
\end{table}

\subsection{Attacking defenses}
We show how QuEry Attack breaks a collection of defense strategies designed to boost the robustness of models against adversarial attacks.

\paragraph{Attacking non-differentiable transformations}
Gradient masking is achieved via non-differentiable input transformations, which rely on manipulating gradients to defeat gradient-based attackers \cite{athalye2018obfuscated,qiu2020mitigating}. Further, randomized transformations make it more difficult for the attacker to be certain of success. It is possible to foil such a defense by altering the defense module that performs gradient masking, but this is not an option within the black-box scenario. Herein, we investigated three non-differentiable transformations against QuEry Attack: JPEG compression, bit-depth reduction (also known as feature squeezing), and spatial smoothing. 
We show that QuEry Attack can defeat these input modifications, due to its gradient-free nature. 

\begin{figure}
    \centering
    \includegraphics[width=0.48\textwidth]{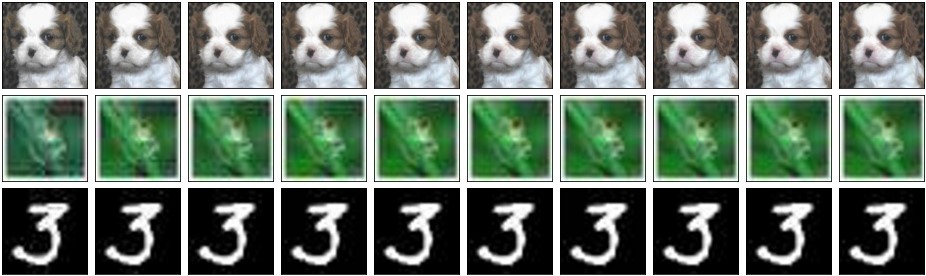}
    \caption{JPEG compression examples, sorted from left to right by quality value, ranging from 10 to 100 (original image). Top images: from ImageNet, middle: CIFAR10, bottom: MNIST.}
    \label{fig:jpeg_compress}
\end{figure}

JPEG compression \cite{dziugaite2016study} tries to generate patterns in color values to minimize the amount of data that has to be captured, resulting in a smaller file size. Some color values are estimated to match those of surrounding pixels in order to produce these patterns. This compression means that slight imperfections in the quality of the image will not be as noticeable. The degree of compression may be tweaked, providing a customizable trade-off between image quality and storage space. An example of the different compression degrees is shown in Figure~\ref{fig:jpeg_compress}. The results in Table~\ref{tab:defense_results} were evaluated with image quality $q=70$.

\begin{figure}
    \centering
    \includegraphics[width=0.48\textwidth]{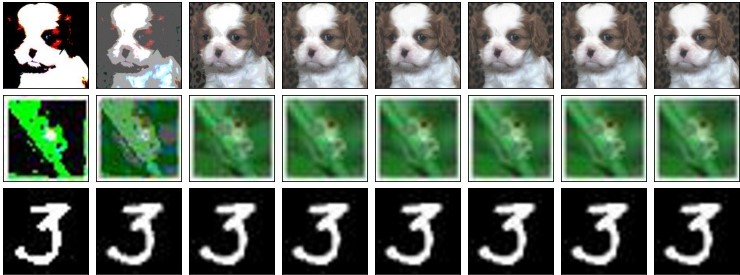}
    \caption{Bit-depth compression examples, sorted from left to right by bit-depth values, ranging from 1 to 8 (original image). Top images: from ImageNet, middle: CIFAR10, bottom: MNIST.}
    \label{fig:bit_depth_compress}
\end{figure}

Bit-depth reduction \cite{xu2017feature} can be done both by reducing the color depth of each pixel in an image and using spatial smoothing to smooth out individual pixel discrepancies. By merging samples that correspond to many different feature vectors in the original space into a single sample, bit-depth reduction decreases the search space accessible to an opponent.  An example of different bit-depth values is shown in Figure~\ref{fig:bit_depth_compress}. The results in Table~\ref{tab:defense_results} were evaluated with bit depth $d=3$.

\begin{figure}
    \centering
    \includegraphics[width=0.48\textwidth]{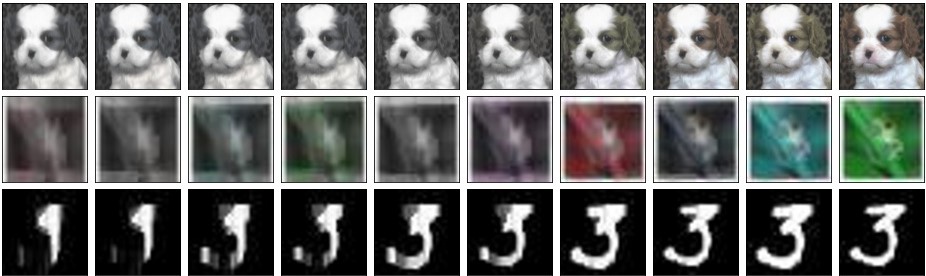}
    \caption{Examples of spatial-smoothing compression, sorted from left to right by window-size values, ranging from 10 to 1 (original image). Top images: from ImageNet, middle: CIFAR10, bottom: MNIST.}
    \label{fig:spas_compress}
\end{figure}

The term ``spatial smoothing`` refers to the averaging of data points with their neighbors \cite{mikl2008effects}. This has the effect of a low-pass filter, with high frequencies of the signal being eliminated from the data while low frequencies are enhanced. As a result, an image's crisp ``edges'' are softened, and spatial correlation within the data becomes more prominent, as shown in Figure~\ref{fig:spas_compress}. Data averaging is determined according to a given window size. The results in Table~\ref{tab:defense_results} were evaluated with window $w=5$.

Our results are delineated in Table~\ref{tab:defense_results}. For this experiment we used a total budget of 82K queries to the model ($N=70$, $G=1200$)---which was Inception-v3. For each given image, we first checked that it was correctly classified after applying the defense on the image, then we applied QuEry Attack on it. The different input values for the transformations were chosen such that applying them would not be destructive.

\begin{table*}
\centering
\small
\caption{QuEry Attack's resistance to non-differentiable transformation defenses. 
         JPEG compression, bit-depth reduction, and spatial smoothing were tested on CIFAR10 and ImageNet.}
\label{tab:defense_results}
\begin{tabular}{|c|c|cc|cc|}
\hline
\multirow{2}{*}{Defense} & \multirow{2}{*}{$\mathcal{E}$} & \multicolumn{2}{c|}{CIFAR10} & \multicolumn{2}{c|}{ImageNet} \\ \cline{3-6} 
                        & &  \multicolumn{1}{l|}{ASR} & Queries & \multicolumn{1}{c|}{ASR} & Queries  \\ \hline

\multirow{3}{*}{JPEG Compression $(q=70)$} & 18 & \multicolumn{1}{c|}{100\%}  & 2 & \multicolumn{1}{c|}{100\%} & 2  \\ 
& 12 & \multicolumn{1}{c|}{98.0\%}  & 13 & \multicolumn{1}{c|}{97.5\%} & 14  \\ 
                             & 6 & \multicolumn{1}{c|}{93.5\%} & 287 & \multicolumn{1}{c|}{99.5\%} & 311  \\ \hline
\multirow{3}{*}{Bit-Depth Reduction $(d=3)$} & 18 & \multicolumn{1}{c|}{100\%}  & 2 & \multicolumn{1}{c|}{100\%} & 2  \\
& 12  & \multicolumn{1}{c|}{99.5\%}  & 5 & \multicolumn{1}{c|}{100\%} &  27  \\ 
                             & 6 &  \multicolumn{1}{c|}{96.0\%}  & 72 & \multicolumn{1}{c|}{99.5\%} & 145  \\
                             \hline
 \multirow{3}{*}{Spatial Smoothing $(w=5)$} & 18 & \multicolumn{1}{c|}{100\%}  & 1 & \multicolumn{1}{c|}{100\%} & 1  \\
 & 12   & \multicolumn{1}{c|}{100\%}  & 1 & \multicolumn{1}{c|}{100\%} &  2  \\ 
 & 6 & \multicolumn{1}{c|}{99.0\%}  & 6 & \multicolumn{1}{c|}{99.5\%} & 144  \\
 \hline

\end{tabular}
\normalsize
\end{table*}

For these experiments we used the same budget of queries as in the previous experiments. For both CIFAR-10 and ImageNet, QuEry Attack has a high success rate against all non-differentiable transformations.

\paragraph{Attacking robust models}
A model is considered to be robust when some of the input variables are largely perturbed, but the model still makes correct predictions.
Recently, several techniques have been proposed to render the models more robust to adversarial attacks. One commonly used technique to improve model robustness is adversarial training. Adversarial training integrates adversarial inputs---generated with other trained models---into the models' training data. Adversarial training has been proved to be one of the most successful defense mechanisms for adversarial attacks \cite{ganin2016domain,tramer2017ensemble,shafahi2019adversarial,wong2020fast}. 

We conducted an experiment to see how well QuEry Attack performs on robust models. For CIFAR10 we used the robust model, WideResNet-70-16 \cite{gowal2021improving}, wherein they used generative models trained only on the original training set in order to enhance adversarial robustness to $l_p$ norm-bounded perturbations. For ImageNet we used WideResNet-50-2 \cite{salman2020adversarially}, wherein they used adversarially robust models for transfer learning. Both of these top models were taken from the RobustBench repository \cite{croce2021robustbench}. We used the same 200 randomly selected images from our previous experiments, a budget of 84K queries ($N=70$, $G=1200$) for CIFAR10, and a budget of 126K queries ($N=70$, $G=1800$) for ImageNet.  As seen in Table~\ref{tab:robust_results}, QuEry Attack succeeds at breaking those strongly defended models. 

\begin{table*}
\centering
\small
\caption{QuEry Attack's performance on state-of-the-art robust models over CIFAR10 and ImageNet.}
\label{tab:robust_results}
\begin{tabular}{|c|c|cc|}
\hline
\multirow{2}{*}{Model} & \multirow{2}{*}{$\mathcal{E}$} & \multicolumn{2}{c|}{ImageNet}  \\ \cline{3-4} 
                        & &  \multicolumn{1}{l|}{ASR} & Queries \\ \hline

\multirow{2}{*}{Wide ResNet-50-2} & 12 & \multicolumn{1}{c|}{98.0\%}  & 98   \\ 
                             & 6 & \multicolumn{1}{c|}{93.5\%} & 1187 \\ \hline
                             
% \multirow{2}{*}{\textbf{Model}} & \multirow{2}{*}{\textbf{Epsilon}} & \multicolumn{2}{c|}{\textbf{CIFAR10}}  \\ \cline{3-4} 
                                % & & \multicolumn{1}{l|}{\textbf{ASR}} & \textbf{ Queries} \\ \hline
\multicolumn{2}{|c|}{} &  \multicolumn{2}{c|}{CIFAR10}  \\ \hline 
\multirow{2}{*}{Wide ResNet-70-16} & 18 & \multicolumn{1}{c|}{94.5\%}  & 156   \\ 
& 12 & \multicolumn{1}{c|}{85.0\%} & 487 \\ \hline
\end{tabular}
\normalsize
\end{table*}

\subsection{Transferability}
An adversarial example for one model can often serve as an adversarial example for another  model, even if the two models were trained on different datasets, using different techniques; this is known as transferability \cite{papernot2016transferability}. White-box attacks may overfit on the source model, as evidenced by the fact that black-box success rates for an attack are almost always lower than those of white-box attacks \cite{carlini2017towards,madry2017towards,brendel2017decision,chen2017zoo}. Herein, we checked transferability of our proposed black-box attack on 200 correctly classified ImageNet images by both the source model and the target model, using different $\epsilon$ values. The results,  summarized in Table~\ref{tab:transfer_results}, show a positive correlation between the $\epsilon$ values and the transferability success rate. We noted that attacks are better transferred between ResNet-50 to VGG-16-BN models and surmise this is due to the fact that ResNets models were mostly inspired by the philosophy of VGG models, wherein they use relatively small $3 \times 3$ convolutional layers.

\begin{table*}
\centering
\small
\caption{QuEry Attack's transferability on ImageNet models. TSR: Transferability Success Rate}
\label{tab:transfer_results}
\begin{tabular}{|c|c|c|}
\hline
% \multirow{2}{*}{Source Model $\rightarrow$ Target Model} & \multirow{2}{*}{Epsilon} & \multicolumn{2}{c|}{ImageNet}  \\ \cline{2-3} 
%                         & &  \multicolumn{1}{l|}{ASR} \\ \hline
\multirow{1}{*}{Source Model $\rightarrow$ Target Model} & \multirow{1}{*}{$\mathcal{E}$} & \multirow{1}{*}{TSR} \\ \hline

\multirow{4}{*}{Inception-v3 $\rightarrow$ ResNet-50} & 24 & \multicolumn{1}{c|}{67.0\%}  \\ 
                             & 18 & \multicolumn{1}{c|}{47.5\%}  \\ 
                             & 12 & \multicolumn{1}{c|}{33.5\%}  \\
                             & 6 & \multicolumn{1}{c|}{13.5\%}  \\ \hline
                             
\multirow{4}{*}{ResNet-50 $\rightarrow$ VGG-16-BN} & 24 & \multicolumn{1}{c|}{90.0\%}    \\ 
                             & 18 & \multicolumn{1}{c|}{81.5\%}  \\ 
                             & 12 & \multicolumn{1}{c|}{61.5\%}  \\ 
                             & 6 & \multicolumn{1}{c|}{28.5\%}  \\ \hline              
                             
\multirow{4}{*}{VGG-16-BN $\rightarrow$ Inception-v3} & 24 & \multicolumn{1}{c|}{59.0\%}   \\ 
                             & 18 & \multicolumn{1}{c|}{46.5\%}  \\ 
                             & 12 & \multicolumn{1}{c|}{30.0\%} \\ 
                             & 6 & \multicolumn{1}{c|}{12.5\%} \\ \hline   
                             
\end{tabular}
\normalsize
\end{table*}

\section{Discussion and Concluding Remarks}
\label{sec:conclusions}
We presented an evolutionary, score-based, black-box attack, showing its superiority in terms of ASR (attack success rate) and number-of-queries over previously published work. QuEry Attack is a strong and fast attack that employs a gradient-free optimization strategy. We tested QuEry Attack against MNIST, CIFAR10, and ImageNet models, comparing it to other state-of-the-art algorithms. We evaluated QuEry Attack's performance against non-differential transformations and robust models, and it proved to succeed in both scenarios.

As noted, we discovered that the software scene in adversarial attacks is a tad bit muddy. We encourage researchers to place executable code on public repositories---code that can be used with ease. Furthermore, we feel that the field lacks standard means of measuring and comparing results. We encourage the community to establish common baselines for these purposes.

We came to realize the importance of a strong initialization procedure.
While this is true of many optimization algorithms, it seems doubly so where adversarial optimization is concerned. Table~\ref{tab:results} shows that successful attacks are sometimes found during initialization---the vertical-stripes initialization in particular proved highly potent---and even if not, the number of queries (and generations) is significantly curtailed.

\begin{figure}
    \centering
    \includegraphics[width=0.48\textwidth]{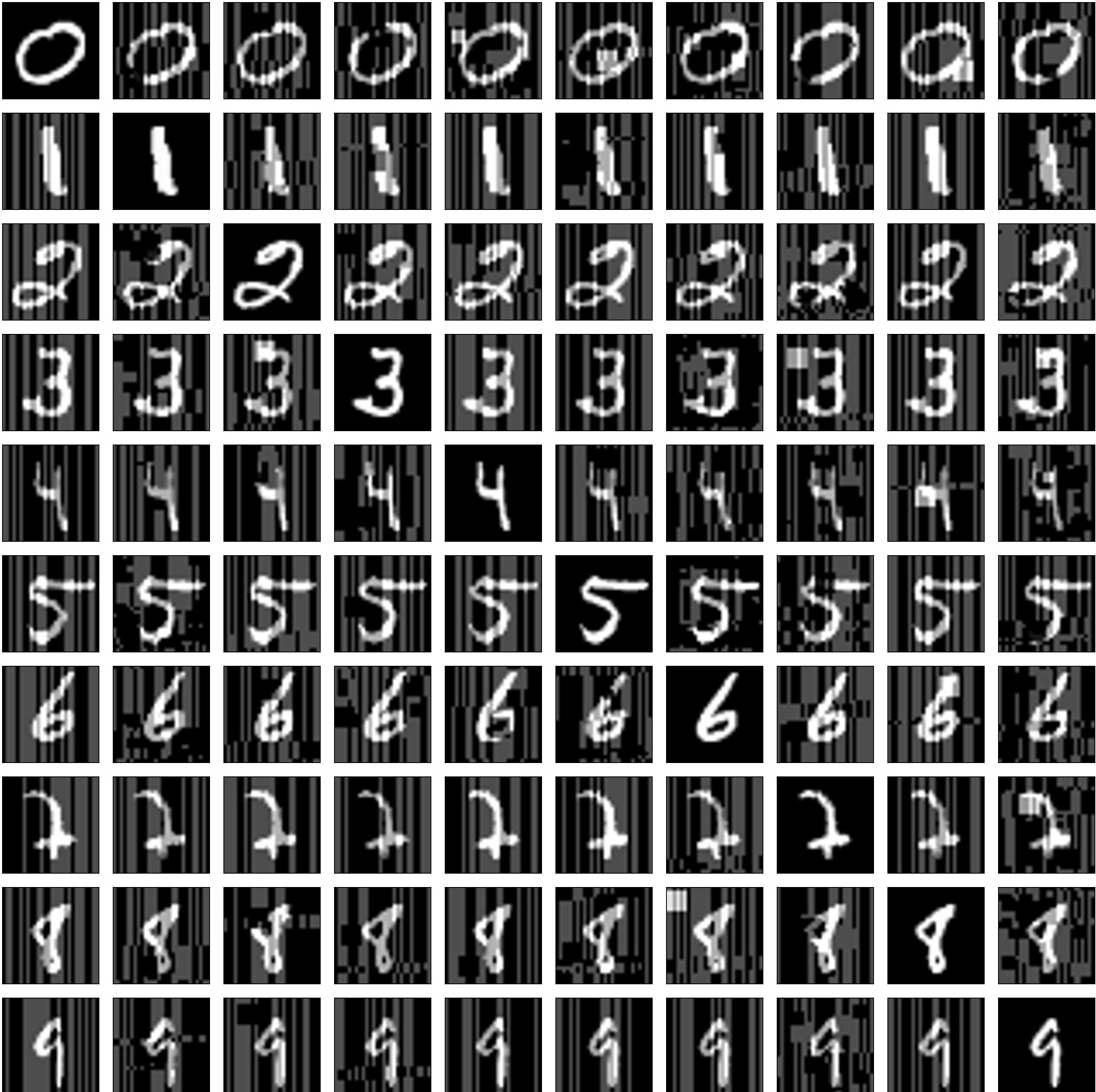}
    \caption{Adversarial examples generated by QuEry Attack on MNIST, with $\epsilon=0.3$. 
             An image at $\mathit{row,col}=i,i$ shows the original image for class $i$.
             An image at $\mathit{row,col}=i,j$, $i \neq j$ shows a targetted attack on class $i$, with the target being class $j$.}
    %The true label is stated in each row, whereas the targeted label is states in the bottom of each column. The original images lie on the digonal.}
    \label{fig:mnist_grid}
\end{figure}

\begin{figure}
    \centering
    \includegraphics[width=0.48\textwidth]{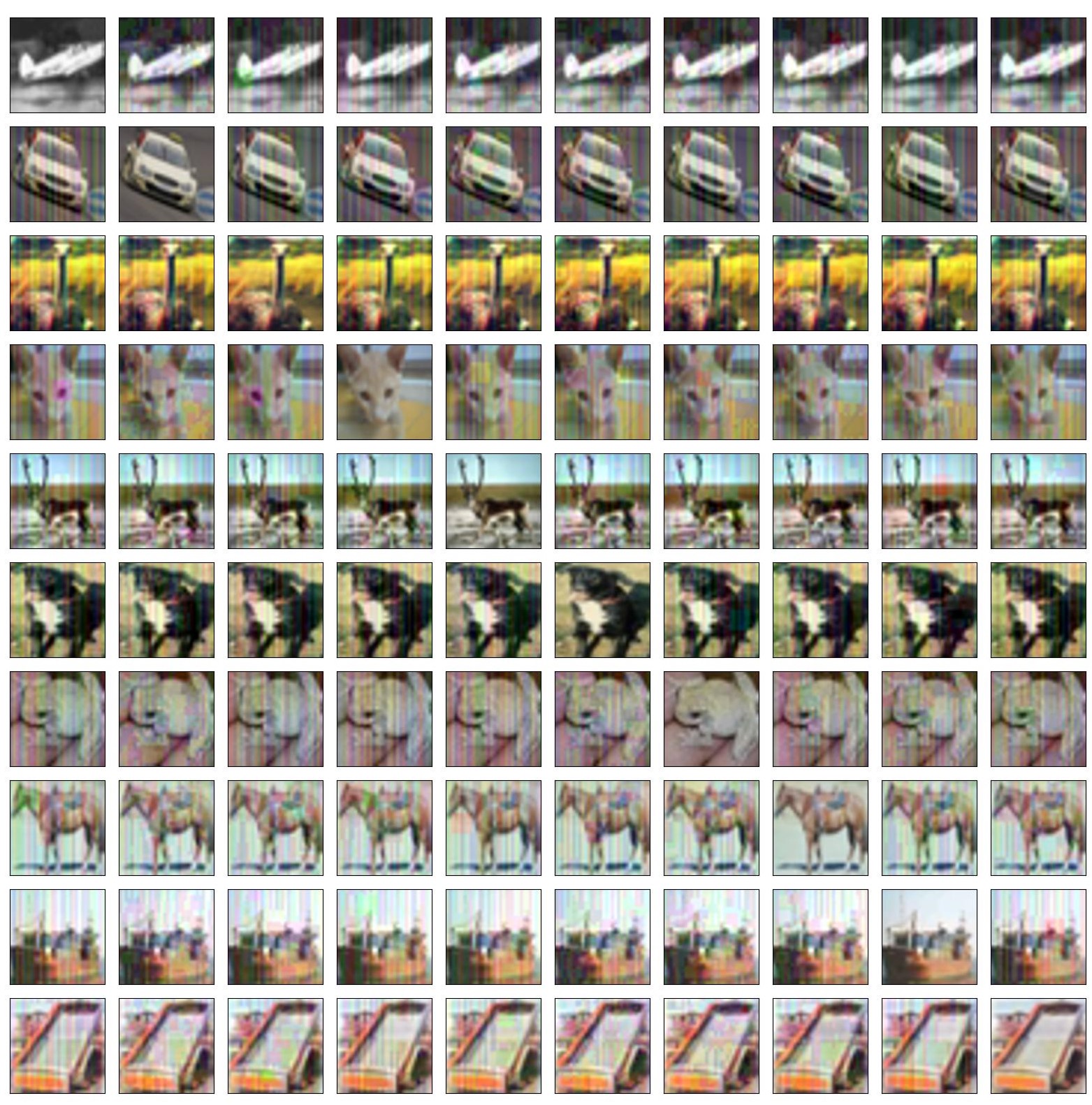}
    \caption{Adversarial examples generated by QuEry Attack on CIFAR10, with $\epsilon=12$. 
             An image at $\mathit{row,col}=i,i$ shows the original image for class $i$.
             An image at $\mathit{row,col}=i,j$, $i \neq j$ shows a targetted attack on class $i$, with the target being class $j$.}
    %The true label is stated in each row, whereas the targeted label is states in the bottom of each column. The original images lie on the digonal.}
    \label{fig:cifar_grid}
\end{figure}

Figures~\ref{fig:mnist_grid} and~\ref{fig:cifar_grid} show that adversarial examples are barely distinguishable to the human eye. Clearly, neural networks function quite differently than humans, capturing entirely different features. More work is needed to create networks that are robust in a human sense.

% While evaluating the attacks over the non-differential transformations, for some of the defense mechanisms we observed better performance over the undefended models---an occurrence we did not expect. In our opinion those techniques need to be inserted as part of the training phase, resulting in additional training instances, just as with adversarial training.

We think that evolutionary algorithms are well-suited for this kind of optimization problems and our findings imply that evolution is a potential research avenue for developing gradient-free black-box attacks. Furthermore, evolution needs to be evaluated against a fully black-box model.

\section*{Acknowledgements}
We thank Ofer Hadar and Itai Dror for helpful discussions. This research was partially supported by the Israeli Innovation Authority and the Trust.AI consortium.

\bibliographystyle{IEEEtran}
\bibliography{refs,Adversarial}

\end{document}